%% file: main.tex
\documentclass[10pt,twocolumn,letterpaper]{article}

\usepackage[pagenumbers]{iccv} 

\input{preamble}

\definecolor{iccvblue}{rgb}{0.21,0.49,0.74}
\usepackage[pagebackref,breaklinks,colorlinks,allcolors=iccvblue]{hyperref}

\def\paperID{} 
\def\confName{}
\def\confYear{}

\title{EVA: Mixture-of-Experts Semantic Variant Alignment for Compositional Zero-Shot Learning}

\author{Xiao Zhang, Yongqiang Ma, Haodong Jing, Nanning Zheng\thanks{Corresponding author.} \\
National Key Laboratory of Human-Machine Hybrid Augmented Intelligence,\\ 
National Engineering Research Center for Visual Information and Applications, \\
and Institute of Artificial Intelligence and Robotics, Xi'an Jiaotong University, Shaanxi, China \\
}

\begin{document}
\maketitle
\input{sec/0_abstract}    
\input{sec/1_intro}
\input{sec/2_relatedwork}
\input{sec/3_method}
\input{sec/4_experiment}
\input{sec/5_conclusion}

{
    \small
    \bibliographystyle{ieeenat_fullname}
    \bibliography{main}
}


\end{document}

%% file: preamble.tex
\usepackage{xcolor}
\usepackage{multirow}
\usepackage{colortbl}

%% file: sec/0_abstract.tex
\begin{abstract}
Compositional Zero-Shot Learning (CZSL) investigates compositional generalization capacity to recognize unknown state-object pairs based on learned primitive concepts. 
Existing CZSL methods typically derive primitives features through a simple composition-prototype mapping, which is suboptimal for a set of individuals that can be divided into distinct semantic subsets. 
Moreover, the all-to-one cross-modal primitives matching neglects compositional divergence within identical states or objects, limiting fine-grained image-composition alignment.
In this study, we propose \textbf{EVA}, a Mixture-of-\underline{E}xperts Semantic \underline{V}ariant \underline{A}lignment framework for CZSL. Specifically, we introduce \textbf{domain-expert adaption}, leveraging multiple experts to achieve token-aware learning and model high-quality primitive representations. To enable accurate compositional generalization, we further present \textbf{semantic variant alignment} to select semantically relevant representation for image-primitives matching. 
Our method significantly outperforms other state-of-the-art CZSL methods on three popular benchmarks in both closed- and open-world settings, demonstrating the efficacy of the proposed insight.
\end{abstract}

\begin{figure}[t]
\centering
\includegraphics[width=\linewidth]{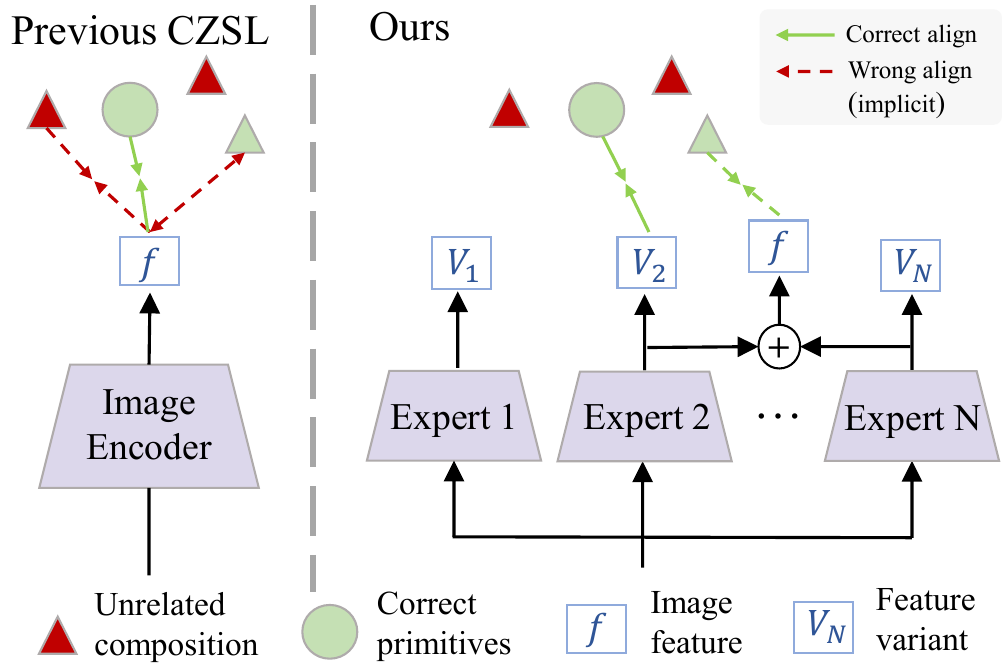}
\caption{Previous primitive alignment forces the composition visual features to be compressed into the primitives space, which neglects compositional divergence within identical primitives and disrupt the topological structure of cross-modal fine-grained associations.
Our method selects most semantically relevant feature variant and fused feature from domain-experts for accurate image-primitives and image-composition alignment.}
\label{fig:ow}
\vspace{-15pt}
\end{figure} 

%% file: sec/1_intro.tex
\vspace{-15pt}
\section{Introduction}
\label{sec:intro}

Humans can identify potential attributes within visual objects and apply learned knowledge to recognize new patterns~\cite{lake2014towards}, such as understanding \textit{blue grass} (even if unseen) based on the concepts of \textit{blue} and \textit{grass}. Compositional generalization~\cite{atzmon2016learning, lake2017building} enables artificial intelligence systems to derive new concepts from existing knowledge, thereby accurately interpreting unseen visual instances. Inspired by this capacity for compositional generalization, Compositional Zero-Shot Learning (CZSL)~\cite{misra2017red, li2020symmetry, naeem2021learning} aims to identify unseen compositions in testing by leveraging state and object knowledge learned during training, facilitating compositional generalization in concept learning.

Existing CZSL solutions~\cite{misra2017red, Huang_2024_CVPR, xu2024gipcol, li2024context, lu2023decomposed, jing2024retrieval} typically leverage state and object information to assist in establishing robust image-composition matching relations. Relying on high-quality features from pre-trained encoders (\eg, CLIP~\cite{radford2021learning}), recent methods~\cite{lu2023decomposed, Huang_2024_CVPR, li2024context, bao2024prompting} achieve impressive performance by introducing well-designed learning strategies, \eg, graph learning~\cite{xu2024gipcol}, cross-modal fusion~\cite{Huang_2024_CVPR}, and prompting language-informed distributions~\cite{xu2022prompting}.

However, there are several limitations.
Firstly, 
previous methods~\cite{li2022siamese, lu2023decomposed, Huang_2024_CVPR, xu2024gipcol} process composition visual features from the image encoder through simple modules (\ie, multi-layer perceptron (MLP)~\cite{misra2017red}) to derive primitives (\ie, state and object) visual features. We argue that this simple mapping is suboptimal for prototype concept learning, as the sematic meanings of different primitives are distinct, and a single ``expert'' cannot effectively capture the semantic patterns of all prototypes. Moreover, inherent information degradation occurs when inferring basic concepts (\ie, state and object) from more specific ones (\ie, composition), since sub-concepts contain only partial information about their larger counterparts. For instances, a \textit{broken car} only contains partial content of a \textit{car}. 

Secondly, as depicted in \cref{fig:ow}, 
prior methods~\cite{purushwalkam2019task, lu2023decomposed, mancini2021open, xu2024gipcol} typically align different composition visual features containing identical state or object with a single primitive text feature. This coarse-grained constraint disrupts the topological structure of cross-modal fine-grained associations and hampers the modeling of discriminative composition features, leading to semantic entanglement in composition representation space. For instance, \textit{blue car}, \textit{blueberry}, and \textit{blue sky} all sharing same attribute \textit{blue}, contain different visual objects, resulting in distinct semantic meanings. Essentially, they neglect the implicit composition feature substructure within the primitive feature cluster.
Consequently, the all-to-one image-primitive alignment confuses fine-grained image-composition generalization.

In light of above limitations, we propose \textbf{EVA}, a Mixture-of-\textbf{\underline{E}}xperts Semantic \textbf{\underline{V}}ariant \textbf{\underline{A}}lignment framework for CZSL. \textbf{EVA} leverages two effective strategies: \textit{domain-expert adaption} for learning high-quality primitives representations, and \textit{semantic variant alignment} for establishing robust fine-grained image-composition alignment.
 
For \textit{domain-expert adaption}, we introduce the Mixture-of-Experts (MoE)~\cite{Jacobs1991AdaptiveMO, Shazeer2017OutrageouslyLN} adapter to process tokens in each layer of the image and text encoders, respectively. Through dynamic token allocation, each expert handles semantically similar tokens, facilitating the mastery of \textbf{in-domain knowledge} and improving performance in prototype semantic modeling.
The MoE adapter learns prototypical concepts at the token level and integrates composition information through the self-attention layer of the pre-trained encoder, ensuring the effective transmission of semantic information. 
To alleviate \textbf{knowledge redundancy} in multi-expert collaboration, we designate a shared expert to capture general knowledge, allowing the other activated experts to focus on specialized knowledge. Compared to suffix modules in \cite{mancini2022learning, bao2024prompting, Huang_2024_CVPR, xu2024gipcol}, \textbf{EVA} is an efficient and flexible \textbf{end-to-end model} for deriving new approaches.

For \textit{semantic variant alignment}, we propose the \textbf{global-to-local} cross-modal alignment from image and text views respectively. In the text domain, since compositions belong to their respective state and object sets, primitive features can be regarded as the centroids of relevant composition features.
Thus, without explicitly maintaining a cluster of feature variants, \textbf{EVA} captures fine-grained primitive features in the text domain. In the image domain, we utilize \texttt{CLS} tokens from various experts as image feature variants, which are semantic representations of visual content from different perspectives. We then measure the similarity between these variants and their corresponding state and object text features respectively, selecting the highest-scoring variants as the primitive visual features.
By considering the differences among various instances under the same primitives, our method achieves more accurate image-state and image-object alignment, effectively guiding zero-shot generalization at the composition level.

To evaluate the proposed method, we conduct comparative experiments on three well-known datasets (\ie, MIT-States~\cite{isola2015discovering}, UT-Zappos~\cite{yu2014fine}, and C-GQA~\cite{naeem2021learning}) in both closed- and open-world settings. Our method significantly outperforms other state-of-the-art CZSL approaches, achieving \textbf{+1.5\%} and \textbf{+4.4\%} AUC gains on MIT-States and C-GQA in closed-world setting. Extensive ablation studies robustly demonstrate the effect of the \textbf{EVA} components.

\begin{figure*}[tb]
\centering
\includegraphics[width=0.92\linewidth]{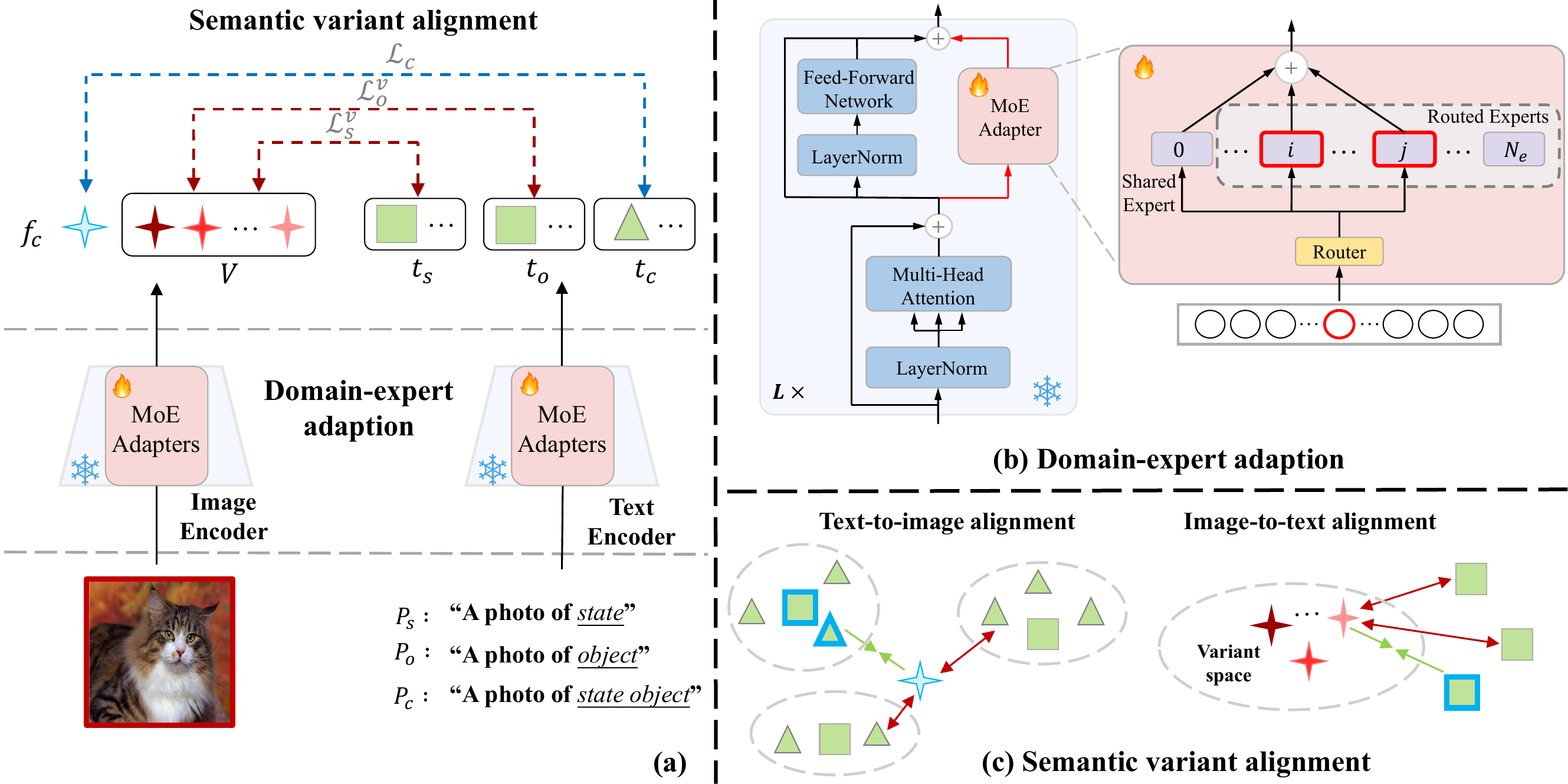}
\caption{(a). The framework of \textbf{EVA} consists of \textit{domain-expert adaption} for token-aware representation learning and \textit{semantic variant alignment} for fine-grained image-primitives matching. (b). \textit{Domain expert adaption} leverages MoE adapter to dynamically process semantically relevant tokens with in-domain knowledge. (c). \textit{Semantic variant alignment} introduces text-to-image and image-to-text alignment to select most relevant features for cross-modal primitives matching from the text and image views, respectively.}
\label{fig:fram}
\vspace{-15pt}
\end{figure*} 

%% file: sec/2_relatedwork.tex
\section{Related Work}
\label{sec:rw}

\textbf{Compositional Zero-shot Learning.} CZSL~\cite{misra2017red, li2020symmetry, lu2023decomposed, nayak2022learning, atzmon2020causal} learns the entanglement between states and objects from seen compositions in training to recognize unseen compositions during test. Recent methods~\cite{nayak2022learning, xu2022prompting, Huang_2024_CVPR, jing2024retrieval} leverage high-quality features from pre-trained Vision-Language Models (VLM)~\cite{radford2021learning}, achieving impressive zero-shot performance. 
CSP~\cite{nayak2022learning} introduces the learnable soft prompt to adapt CLIP~\cite{radford2021learning} to CZSL tasks. DFSP~\cite{lu2023decomposed} designs a cross-modal decomposed fusion to make image features more expressive for image-primitive alignment. Troika~\cite{Huang_2024_CVPR} proposes a cross-modal traction module to adaptively learn text representations relevant to visual content. However, they utilize single ``expert'' module for various tokens with different semantic contents and fail to achieve deep primitives learning relying on suffix modules. In this study, we propose \textit{domain-expert adaption} to introduce MoE~\cite{Shazeer2017OutrageouslyLN} adapters at each layer of encoders, enabling deep and efficient in-domain primitive feature learning.

\noindent \textbf{Concept Representation Learning.}
Learning transferable and generalizable concept representations is a central challenge in deep learning. In natural language processing~\cite{vaswani2017attention, mikolov2013efficient}, several methods~\cite{Devlin2019BERTPO, radford2019language, brown2020language, dong2019unified} attempt to pretrain models on large-scale datasets, learning general and accurate token representations. The impressive performance of large language models (LLM)~\cite{achiam2023gpt, team2024gemini} validates the effectiveness of this paradigm. Recent works ~\cite{jiang2024mixtralexperts, liu2024deepseek} further utilize the Mixture-of-Experts (MoE) layer to achieve dynamic token routing for in-domain knowledge learning. However, identifying fundamental visual concepts for description remains challenging. Visual concept learning typically requires natural language supervision~\cite{radford2021learning, li2023blip, liu2024visual, li2021align}. CLIP~\cite{radford2021learning} learns transferable visual representations through contrastive learning on large-scale image-pair datasets. BLIP~\cite{li2022blip} is pre-trained with both understanding-based and generation-based tasks to learn generalizable representations. LLaVA~\cite{liu2024visual} introduces visual instruction tuning to align LLM with the multimodal space. In this work, we propose a multi-expert representation method tailored for the semantic modeling of concepts, enhancing the expressiveness of primitive features for cross-modal alignment.

%% file: sec/3_method.tex
\vspace{-7pt}
\section{Methodology}
\subsection{Preliminary}

The goal of CZSL is to develop a model with compositional generalization capacity, enabling it to accurately recognize unseen compositions based on learned primitive knowledge (\ie, state and object).
Given the state label set $\mathcal{S} = \{ s_1, s_2, \dots, s_m \}$ and the object label set $\mathcal{O} = \{ o_1, o_2, \dots, o_n \}$, the state-object composition set $\mathcal{C}$ is defined as the Cartesian product of these two primitive concepts sets (\ie, $\mathcal{C} = \mathcal{S} \times \mathcal{O}$). For zero-shot evaluation, we denote seen and unseen composition label sets as $\mathcal{C}^{s}$ and $\mathcal{C}^{u}$ respectively, which are disjoint subsets of $C$. The training dataset is defined as $\mathcal{T} = \{ (x_i, c_i) | x_i \in \mathcal{X}, c_i \in \mathcal{C}^{s} \}$, where $\mathcal{X}$ denotes the image space, and only the seen set $\mathcal{C}^{s}$ is accessible during training. In closed-world setting~\cite{naeem2021learning}, the test composition set $\mathcal{C}^{t}$ includes both seen and unseen sets, defined as $\mathcal{C}^{t} = \mathcal{C}^{s} \cup \mathcal{C}^{u}$. In the more challenging open-world setting~\cite{mancini2021open}, the state-object composition set $\mathcal{C}$ is utilized as the test composition space.  

\subsection{Framework Overview}
The human brain consists of different functional areas that work together to complete various tasks. Inspired by distributed functional system, we propose \textbf{EVA}, a Mixture-of-\textbf{\underline{E}}xpert Semantic \textbf{\underline{V}}ariant \textbf{\underline{A}}lignment framework for CZSL, which employs several domain-experts for adaptive concept learning and fine-grained semantic alignment, as depicted in \cref{fig:fram} (a).
Specifically, we introduce \textit{domain-expert adaption}, illustrated in \cref{fig:fram} (b), to dynamically process tokens in each
layer of the image and text encoders, respectively. The proposed domain-experts create a sophisticated mapping to deeply learn primitives at the token level.
To address existing all-to-one image-primitives alignment, we design \textit{semantic variant alignment}, shown in \cref{fig:fram} (c), to select the most semantically relevant features variants for fine-grained primitive concept learning.

\subsection{Domain-Expert Adaption}
In this study, we adopt the frozen CLIP~\cite{radford2021learning} image encoder $E_v$ and text encoder $E_t$ to derive image and text representations, respectively. Unlike previous methods~\cite{lu2023decomposed, Huang_2024_CVPR, misra2017red, mancini2022learning}, \textbf{EVA} introduces MoE intra-layer adapters, making it an efficient end-to-end model that does not require suffix modules.

As depicted in \cref{fig:fram} (b), the MoE adapter, parallel to the Feed-Forward Network (FFN), consists of a router $\mathcal{R}$ for dynamic token allocation and multiple experts $\{\mathcal{E}_i\}_{i=0}^{N_E}$ for domain-aware representation learning. 
It is worth noting that we designate a shared expert $\mathcal{E}_0$ to learn common knowledge, while others serve as routed experts $\{\mathcal{E}_i\}_{i=1}^{N_{E}}$ to focus on domain-specific knowledge. 
We utilize LoRA~\cite{Hu2021LoRALA} method to develop the MoE adapters, where each expert $\mathcal{E}_i$ has an identical structure with two trainable parameters $A \in \mathbb{R}^{r \times d}$, $B \in \mathbb{R}^{d \times r}$ and $r \ll d$.
Given the hidden token embedding $h_j \in \mathbb{R}^{d}$ in layer $j$, the router first computes the token-to-expert affinity $G \in \mathbb{R}^{k}$ to measure each expert’s contribution. Then each token is assigned to shared expert and $K$ routed experts for domain-knowledge learning:
\begin{align}
    & G = Softmax(TopK(\mathcal{R}(h_j))), \\
    & \mathcal{E}_i = B_iA_i, \\
    & h_{j+1} = \sum_{i=1}^{N_e} G_i\mathcal{E}_i(h_j) + \mathcal{E}_0(h_j),
\end{align} 
where $Topk(\cdot)$ function selects the $K$ most relevant experts, while setting  the scores of the other experts to $-\infty$.
With the implementation of the MoE adapter in both the image and text encoders, our method leverages domain-expert knowledge and inter-expert collaboration to model high-quality image representation $f_c \in \mathbb{R}^{1 \times d}$ and text representation $t_c \in \mathbb{R}^{1 \times d}$, respectively:
\begin{equation}
    f_c = E_v(x),\ t_c = E_t(P_{c}),
\end{equation}
where $x \in \mathbb{R}^{h \times w \times 3}$ denotes the input image, and $P_c = [\ \theta_1, ..., \theta_a, \theta_s, \theta_o\ ]$ is the learnable composition prompt, initialized with "a photo of state object". Finally, we compute the composition probability corresponding to the image $x$ as follows:
\begin{equation}
    p_{c}(c | x)=\frac{\exp(f_c \cdot t_c^\top / \tau)}{\sum_{y_c \in \mathcal{C}^{S}}\exp(f_c \cdot t_{y_c}^\top / \tau)},
\end{equation}
where $\tau \in \mathbb{R}$ is the temperature coefficient from pre-trained CLIP.
The training objective for composition classification is defined as:
\begin{equation}
\mathcal{L}_{c} = -\frac{1}{|\mathcal{T}|}\sum_{(x,c)\in \mathcal{T}}{\log p_{c}(c|x)}.
\end{equation}

\begin{table*}[t]
\centering
\resizebox{\textwidth}{!}{
\begin{tabular}{l|cccc|cccc|cccc}
\toprule
\multirow{2}{*}{\textbf{Method}} & \multicolumn{4}{c|}{\textbf{MIT-States}} & \multicolumn{4}{c|}{\textbf{UT-Zappos}} & \multicolumn{4}{c}{\textbf{C-GQA}} \\ 
& Unseen $\uparrow$ & Seen $\uparrow$    & AUC $\uparrow$     & HM $\uparrow$  & Unseen $\uparrow$    & Seen $\uparrow$    & AUC $\uparrow$     & HM $\uparrow$  & Unseen $\uparrow$   & Seen $\uparrow$  & AUC $\uparrow$   & HM $\uparrow$ \\ \cmidrule{1-13}
CLIP~\cite{radford2021learning} {\scriptsize \textcolor{gray}{ICML'21}} &46.0 &30.2 &11.0 &26.1 &49.1 &15.8 &5.0 &15.6 &25.0 &7.5 &1.4 &8.6 \\
CoOp~\cite{zhou2022learning} {\scriptsize \textcolor{gray}{IJCV'22}} &47.6 &34.4 &13.5 &29.8 &49.3 &52.1 &18.8 &34.6 &26.8 &20.5 &4.4 &17.1 \\
PromptCompVL~\cite{xu2022prompting} {\scriptsize \textcolor{gray}{arXiv'22}} & 47.2   & 48.5  & 18.3  & 35.3 & 64.0  & 64.4  & 32.2  & 46.1 & - & - & - & - \\
CSP~\cite{nayak2022learning} {\scriptsize \textcolor{gray}{ICLR'23}}  & 49.9   & 46.6  & 19.4  & 36.3  & 66.2  & 64.2  & 33.0  & 46.6 & 26.8 & 28.8 & 6.2 & 20.5 \\
DFSP(\textit{i2t})~\cite{lu2023decomposed} {\scriptsize \textcolor{gray}{CVPR'23}} & 52.4   & 47.4  & 20.7  & 37.2 & 66.4 & 64.2 & 32.1  & 45.1 & 29.3 & 35.6 & 8.7 & 24.3 \\
DFSP(\textit{BiF})~\cite{lu2023decomposed} {\scriptsize \textcolor{gray}{CVPR'23}}  & 52.8   & 47.1  & 20.8  & 37.7 & 69.2 & 63.3 & 33.5  & 47.1 & 32.0 & 36.5 & 9.9 & 26.2 \\
DFSP(\textit{t2i})~\cite{lu2023decomposed} {\scriptsize \textcolor{gray}{CVPR'23}} & 52.0   & 46.9  & 20.6  & 37.3 & 71.7 & 66.7 & 36.0  & 47.2 & 32.0 & 38.2 & 10.5 & 27.1 \\
GIPCOL~\cite{xu2024gipcol} {\scriptsize \textcolor{gray}{WACV'24}} &49.6  &48.5  & 19.9  & 36.6  &68.5  &65.0 & 36.2  & 48.8  & 28.4 & 31.9 & 7.1 & 22.5 \\
CDS-CZSL~\cite{li2024context} {\scriptsize \textcolor{gray}{CVPR'24}} & 52.9 &50.3 &22.4 &39.2 &74.8 &63.9 &39.5 &52.7 &34.2 &38.3 &11.1 &28.1\\
Troika~\cite{Huang_2024_CVPR} {\scriptsize \textcolor{gray}{CVPR'24}} & 53.0 &49.0 &22.1 &39.3 &73.8 &66.8 &41.7 &54.6 &35.7 &41.0 &12.4 &29.4 \\
PLID~\cite{bao2024prompting} {\scriptsize \textcolor{gray}{ECCV'24}} &52.4 &49.7 &22.1 &39.0 &68.8 &67.3 &38.7 &52.4 &33.0 &38.8 &11.0 &27.9\\
RAPR~\cite{jing2024retrieval} {\scriptsize \textcolor{gray}{AAAI'24}} &53.3 &50.0 &22.5 &39.2 &72.8 &69.4 &44.5 &56.5 &36.0 &45.6 &14.4 &32.0\\
\rowcolor[rgb]{.949, .949, .949} \textbf{Ours} & \textbf{55.0} & \textbf{51.2} & \textbf{24.0} & \textbf{41.0} & \textbf{79.6}  & \textbf{71.2} & \textbf{50.2} & \textbf{60.2} & \textbf{44.6} & \textbf{47.1} & \textbf{18.8} & \textbf{36.9} \\ \bottomrule
\end{tabular}}
\caption{\textbf{Quantitative results} on MIT-States, UT-Zappos, and C-GQA in \textbf{Closed-World} setting.}
\label{tab:closed-world}
\end{table*}

\subsection{Semantic Variant Alignment}
Primitive concept (\ie, state and object) learning is essential for establishing robust and accurate image-composition alignment. Existing methods~\cite{misra2017red, mancini2021open, lu2023decomposed, Huang_2024_CVPR} typically align the primitives visual features with state and object text representations, respectively. However, since composition can be seen as a cluster of semantic variants centered on primitives, this all-to-one alignment neglects composition divergence within identical primitives, hindering fine-grained composition matching. To address this limitation, as shown in \cref{fig:fram} (c), we propose \textit{semantic variant alignment} which constructs multiple feature variants for adaptive image-primitives relations. This approach performs a global-to-local cross-modal alignment from both image and text perspectives, respectively.

\noindent \textbf{Text-to-image alignment.} The proposed variant-based method leverages the local composition distribution to select the most semantically relevant individual, rather than original primitives feature. For each state $s$, we select the highest matching score among all compositions within this state as the image-state matching score $P_s$, formulated as:
\begin{equation}
    p_{s}(s | x) = \mathop{\max}_{c_{s, o}}p_{c}(c_{s,o} | x) \cdot \tau_s,
\end{equation}
where $c_{s,o} \in C^{target}$ denotes the state $s$-relevant composition and $C^{target}$ is the target composition label space. $\tau_s > 0$ is a trainable  coefficient used to adjust the distribution of state probabilities. Similarly, given the object $o$, the image-object matching probability $p_o$ can be obtained as follows:
\begin{equation}
    p_{o}(o | x) = \mathop{\max}_{c_{s, o}}p_{c}(c_{s,o} | x) \cdot \tau_o,
\end{equation}
where $\tau_o > 0$ is a trainable coefficient.
To recognize the states and objects, two cross-entropy losses are employed:
\begin{align}
  &\mathcal{L}_{s} = -\frac{1}{|\mathcal{T}|}\sum_{(x,c)\in \mathcal{T}}{\log p_{s}(s|x)}, \\
  &\mathcal{L}_{o} = -\frac{1}{|\mathcal{T}|}\sum_{(x,c)\in \mathcal{T}}{\log p_{o}(o|x)}. 
\end{align}

\noindent \textbf{Image-to-text alignment.} 
Although text-to-image alignment utilizes detailed composition information to mitigate the impact of centralization, the composition-based approach also exacerbates the gap between the seen and unseen sets.
Therefore, we propose bridging this gap by utilizing constant state and object information to refine the well-structured image representation space during training.

Specifically, given a training image $x$ with state label $s$ and object label $o$, we first model the image feature variants $\{ v_i \}_{i=0}^{N_e} \in \mathbb{R}^{N_e \times d}$ from the MoE adapter in the final layer of the image encoder $E_v$. The state feature $t_s \in \mathbb{R}^{d}$ and object feature $t_o \in \mathbb{R}^{d}$ are derived from the text encoder $E_t$ based on learnable prompt $P_s$ and $P_o$:
\begin{equation}
    t_s = E_t(P_s),\ t_o = E_t(P_o),
\end{equation}
where $P_s = [\ \theta_1, ..., \theta_a, \theta_s\ ]$ is initialized with "a photo of state", and $P_o = [\ \theta_1, ..., \theta_a, \theta_o \ ]$ is initialized with "a photo of object".
Since multiple experts exploit semantic information from various representation subspaces, the output representation can be considered as the feature variants that describe different semantic contents of the input image.
Similar to text-to-image alignment, we propose \textit{inter- and intra-model affinity} to select the most semantically relevant variant as the primitive feature.

In terms of \textit{inter-model affinity}, we measure the state affinity score $A_s \in \mathbb{R}^{N_e+1}$ and object affinity score $A_o \in \mathbb{R}^{N_e+1}$ for feature variants $V = \{ v_i \}_{i=0}^{N_e}$ and the corresponding text primitive features, respectively (\ie, $A_s = Vt_s^\top$ and $A_o = Vt_o^\top$). In terms of \textit{intra-model affinity}, we compute the affinity scores $A_v \in \mathbb{R}^{N_e+1}$ for image features and semantic variants, which introduce supervision for the global semantic content (\ie, $A_v = Vf_c^\top$). The affinity score $A_v$ is beneficial for excluding semantic variants that differ significantly from the main semantic content. Furthermore, the overall affinity score $A_S$ and $A_O$ are derived by considering both affinity scores comprehensively:
\begin{equation}
    A_S = A_s + \alpha A_v,\
    A_O = A_o + \alpha A_v,
\end{equation}
where $\alpha > 0$ is a balancing coefficient. Finally, we select the image state feature $f_s$ and object feature $f_o$ from semantic variants $\{ v_i \}_{i=0}^{N_e}$ based on the overall affinity scores:
\begin{align}
    f_s = \mathop{\arg\max}_{v_i} a_i^s,\ \{ a_i^s \in A_S | 0<=i<=N_e\}, \\
    f_o = \mathop{\arg\max}_{v_i} a_i^o,\ \{ a_i^o \in A_O | 0<=i<=N_e\}.
\end{align}
The image-to-text primitives probability is defined as:
\begin{align}
    p_{s}^v(s_i | x)=\frac{\exp(f_s \cdot t_{s}^\top / \tau)}{\sum_{s \in \mathcal{S}}\exp(f_s \cdot t_{s}^\top / \tau)}, \\
    p_{o}^v(o_i | x)=\frac{\exp(f_o \cdot t_{o}^\top / \tau)}{\sum_{o \in \mathcal{O}}\exp(f_o \cdot t_{o}^\top / \tau)},
\end{align}
where $p_{s}^v$ denotes image-to-text state probability and $p_{o}^v$ denotes image-to-text object probability. The training objective of \textit{semantic variant alignment} is defined as:
\begin{align}
    &\mathcal{L}_{s}^v = -\frac{1}{|\mathcal{T}|}\sum_{(x,c)\in \mathcal{T}}{\log p_{s}^v(s|x)}, \\
    &\mathcal{L}_{o}^v = -\frac{1}{|\mathcal{T}|}\sum_{(x,c)\in \mathcal{T}}{\log p_{o}^v(o|x)}. 
\end{align}

\begin{table}[t]
\vspace{-5pt}
\centering
\resizebox{\linewidth}{!}{
\begin{tabular}{lccc|c|cc|cc}
\toprule
     \multirow{2}{*}{\large{Dataset}} &&&& \multicolumn{1}{c|}{Train} & \multicolumn{2}{c|}{Validation} & \multicolumn{2}{c}{Test} \\
     & $|S|$ & $|O|$ &$|C|$ & $|C^s|$ & $|C^s|$ & $|C^u|$ & $|C^s|$ & $|C^u|$ \\
    \midrule
    MIT-States~\cite{isola2015discovering} & 115 & 245 & 28175 & 1262 & 300 &300 &400 &400\\
    UT-Zappos~\cite{yu2014fine} &16 &12 &192 &83 &15 &15 &18 &18 \\
    C-GQA~\cite{naeem2021learning} &413 &674 &278362 &5592 &1252 &1040 &888 &923 \\
\bottomrule
\end{tabular}
}
\vspace{-5pt}
\caption{The detailed data split statistics.}
\label{tab:split}
\end{table}

\subsection{Training and Inference}
\textbf{Training objectives.} The final training objective of \textbf{EVA} is achieved by optimizing all classification loss functions, formulated as:
\begin{equation}
    \mathcal{L} = \mathcal{L}_{c} + \lambda_1 (\mathcal{L}_{s} + \mathcal{L}_{o}) + \lambda_2(\mathcal{L}_s^v + \mathcal{L}_o^v),
\end{equation}
where $\lambda_1$ and $\lambda_2$ are two coefficients. Due to the reliance on label information, \textit{image-to-text alignment} is only applicable during training.

\textbf{Inference.} The final composition prediction $\hat{c}_{s,o}$ combines state $p_s$, object $p_o$, and composition scores $p_c$:
\begin{equation}
    \hat{c}_{s,o} = \mathop{\arg\max}_{c_{s,o}\in \mathcal{C}^{test}} p_{c}(c_{s,o}| x) + \beta (p_s(s|x) + p_o(o|x)),
\end{equation}
where the coefficient $\beta$ is set as $0.5$.

%% file: sec/4_experiment.tex
\begin{table*}[t]
\centering
\resizebox{\textwidth}{!}{
\begin{tabular}{l|cccc|cccc|cccc}
\toprule
\multirow{2}{*}{\textbf{Method}} & \multicolumn{4}{c|}{\textbf{MIT-States}} & \multicolumn{4}{c|}{\textbf{UT-Zappos}} & \multicolumn{4}{c}{\textbf{C-GQA}} \\ 
& Unseen $\uparrow$     & Seen $\uparrow$    & AUC $\uparrow$    & HM $\uparrow$  & Unseen $\uparrow$    & Seen $\uparrow$    & AUC $\uparrow$    & HM $\uparrow$  & Unseen $\uparrow$    & Seen $\uparrow$   & AUC $\uparrow$   & HM $\uparrow$ \\ \cmidrule{1-13}
CLIP~\cite{radford2021learning} {\scriptsize \textcolor{gray}{ICML'21}} &14.3 &30.1 &3.0 &12.8 &20.6 &15.7 &2.2 &11.2 &4.6 &7.5 &0.3 &4.0 \\
CoOp~\cite{zhou2022learning} {\scriptsize \textcolor{gray}{IJCV'22}} &9.3 &34.6 &2.8 &12.3 &31.5 &52.1 &13.2 &28.9 &4.6 &21.0 &0.7 &5.5 \\
PromptCompVL~\cite{xu2022prompting} {\scriptsize \textcolor{gray}{arXiv'22}} & 16.0   & 48.5  & 6.1  & 17.7 & 44.0  & 64.6  & 21.6  & 37.1  & - & - & - & - \\
CSP~\cite{nayak2022learning} {\scriptsize \textcolor{gray}{ICLR'23}} &15.7 &46.3 &5.7 &17.4 &44.1 &64.1 &22.7 &38.9 &5.2 &28.7 &1.2 &6.9 \\
DFSP(\textit{i2t})~\cite{lu2023decomposed} {\scriptsize \textcolor{gray}{CVPR'23}} &18.2 &47.2 &6.7 &19.1 &53.8 &64.3 &26.4 &41.2 &6.5 &35.6 &2.0 &9.0 \\
DFSP(\textit{BiF})~\cite{lu2023decomposed} {\scriptsize \textcolor{gray}{CVPR'23}} &18.1 &47.1 &6.7 &19.2 &57.2 &63.5 &27.6 &42.7 &7.6 &36.4 &2.4 &10.6 \\
DFSP(\textit{t2i})~\cite{lu2023decomposed} {\scriptsize \textcolor{gray}{CVPR'23}} & 18.5   & 47.5  & 6.8  & 19.3  & 60.0 & 66.8  & 30.3  & 44.0  & 7.2 & 38.3 & 2.4 & 10.4 \\
GIPCOL \cite{xu2024gipcol} {\scriptsize \textcolor{gray}{WACV'24}} & 16.0   & 48.5  & 6.3  & 17.9  & 45.0 & 65.0  & 23.5  & 40.1 & 5.5 & 31.6 & 1.3 & 7.3 \\
CDS-CZSL~\cite{li2024context} {\scriptsize \textcolor{gray}{CVPR'24}} &21.8 &49.4 &8.5 &22.1 &61.3 &64.7 &32.3 &48.2 &8.2 &37.6 &2.7 &11.6 \\
Troika~\cite{Huang_2024_CVPR} {\scriptsize \textcolor{gray}{CVPR'24}} &18.7 &48.8 &7.2 &20.1 &61.2 &66.4 &33.0 &47.8 &7.9 &40.8 &2.7 &10.9 \\
PLID~\cite{bao2024prompting} {\scriptsize \textcolor{gray}{ECCV'24}} &18.7 &49.1 &7.3 &20.0 &55.5 &67.6 &30.8 &46.6 &7.5 &39.1 &2.5 &10.6 \\
RAPR~\cite{jing2024retrieval} {\scriptsize \textcolor{gray}{AAAI'24}} &20.1 &49.9 &8.2 &21.8 &59.4 &69.4 &33.3 &47.9 &11.2 &45.5 &4.4 &14.6\\
\rowcolor[rgb]{ .949, .949, .949} \textbf{Ours} & \textbf{23.2} & \textbf{50.8} & \textbf{9.4} & \textbf{22.8} & \textbf{66.5} & \textbf{71.6} & \textbf{40.2} & \textbf{54.2} & \textbf{13.3} & \textbf{46.9} & \textbf{5.6} & \textbf{17.9} \\ \bottomrule
\end{tabular}}
\caption{\textbf{Quantitative results} on MIT-States, UT-Zappos, and C-GQA in \textbf{Open-World} setting.}
\label{tab:open-world}
\vspace{-10pt}
\end{table*}

\section{Experiment}
\subsection{Experiment Setup}
\noindent \textbf{Datasets.} We evaluate the proposed \textbf{EVA} on three widely-used CZSL datasets: MIT-States~\cite{isola2015discovering}, UT-Zappos~\cite{yu2014fine}, and C-GQA~\cite{naeem2021learning}.
MIT-States comprises 53,753 natural images annotated with 115 states, 245 objects, and 1,962 state-object compositions.
UT-Zappos contains 50,025 shoes images with 116 fine-grained compositions, including 16 states and 12 objects.
C-GQA is the largest CZSL dataset, featuring 39,298 images labeled with 7,767 compositions, encompassing 453 states and 870 objects.
In closed-world setting, we follow previous works~\cite{naeem2021learning, lu2023decomposed} to partition the datasets into seen and unseen sets.
The detailed statistics of the dataset splits are provided \cref{tab:split}.

\noindent \textbf{Evaluation Metrics.} Following the evaluation protocol established in prior works~\cite{misra2017red, lu2023decomposed, Huang_2024_CVPR}, we adjust a calibration bias applied to unseen scores from $-\infty$ to $+\infty$ , balancing the prediction scores between seen and unseen pairs. We report the Area Under the Curve (\textbf{AUC}) and the best Harmonic Mean (\textbf{HM}) to quantify the overall performance across seen and unseen compositions. Moreover, we record the best-seen accuracy \textbf{Seen} and best-unseen accuracy \textbf{Unseen} to assess performance on these two disjoint subsets.

\noindent \textbf{Implementation Details.} \textbf{EVA} is built upon the pre-trained frozen CLIP~\cite{radford2021learning} ViT-L/14 model, which serves as both the image and text encoder. In the \textit{domain-expert adaption} stage, the MoE adapter with a LoRA-based~\cite{Hu2021LoRALA} intra-layer structure comprises a router $\mathcal{R}$, a shared expert $\mathcal{E}_0$ and routed experts $\{ \mathcal{E} \}_{i=1}^{N_e}$. The router $\mathcal{R}$ is a single-layer fully connected network, while the shared and routed experts are two-layer MLPs with identical structures, where the hidden dim $r$ is set to 64. The number of activated experts $K$ is set to 2. Moreover, the coefficient
$\alpha$ is set to 0.5 in intra-model affinity. In the final loss function, the weighting parameters $\lambda_1$ and $\lambda_2$ are set to 0.5 and 0.1, respectively.

\noindent \textbf{Training and Test.} We train \textbf{EVA}, implemented in PyTorch~\cite{paszke2019pytorch}, using Adam optimizer~\cite{kingma2014adam} for 20 epochs with a learning rate of $1e-4$. The weight decay is set to $1e-4$, $5e-4$ and $1e-4$ for MIT-States, UT-Zappos, and C-GQA, respectively. Following prior works~\cite{naeem2021learning,lu2023decomposed}, we apply post-training calibration to filter out infeasible compositions in the open-world setting during testing.

\subsection{Comparisons with SoTAs}
We evaluate the quantitative performance of \textbf{EVA} in comparison to previous CLIP-based CZSL methods~\cite{radford2021learning, zhou2022learning, xu2022prompting, nayak2022learning, lu2023decomposed, xu2024gipcol, Huang_2024_CVPR, li2024context, bao2024prompting, jing2024retrieval} in both closed- and open-world settings, as presented in Table \ref{tab:closed-world} and Table \ref{tab:open-world}. 

\noindent \textbf{Evaluation in Closed-World Setting.} As shown in \cref{tab:closed-world}, in closed-world setting, \textbf{EVA} achieves remarkable improvements over other state-of-the-art methods, with AUC gains of $+1.5\%$, $+5.7\%$, and $+4.4\%$ on MIT-States, UT-Zappos and C-GQA, respectively. Specifically, our \textbf{EVA} improves HM by at least $+1.4\%$ compared to previous CZSL methods on MIT-States. Similarly, on UT-Zappos dataset, our method attains an Unseen score of $79.6\%$, surpassing RAPR~\cite{jing2024retrieval} by a significant margin of $6.4\%$, and achieving a $+3.7\%$ improvement in HM. Notably, our method achieves the highest HM score of $36.9\%$ in C-GQA, demonstrating enhanced compositional generalization compared to other methods~\cite{Huang_2024_CVPR, li2024context, jing2024retrieval}.
These results underscore the advantage of domain-expert adaption in token-aware concept representation learning.
In general, our approach \textbf{EVA} exhibits robust performance across all datasets and evaluation metrics, consistently outperforming previous state-of-the-art methods, particularly in unseen sets, underscoring its potential for practical applications in closed-world settings.

\begin{table*}[t]
\centering
\vspace{-5pt}
\resizebox{0.8\textwidth}{!}{
\begin{tabular}{l|cccc|cccc}
\toprule
 & \multicolumn{4}{c|}{\textbf{MIT-States}} & \multicolumn{4}{c}{\textbf{C-GQA}} \\ 
& Unseen $\uparrow$     & Seen $\uparrow$    & AUC $\uparrow$    & HM $\uparrow$  & Unseen $\uparrow$    & Seen $\uparrow$    & AUC $\uparrow$    & HM $\uparrow$ \\ \cmidrule{1-9}
\textsc{Baseline} &51.8 &47.1 &20.2 &36.9 &32.5 &38.3 &10.4 &26.9 \\
Domain-expert Adaption &54.8 &50.1 &23.0 &40.1 &42.8 &45.6 &17.2 &35.5 \\
Semantic Variant Alignment &52.3 &49.8  &22.2 &39.7 &34.7  &41.2  &12.1  &29.8 \\
\rowcolor[rgb]{ .949, .949, .949} Adaption + Alignment &\textbf{55.0} &\textbf{51.2} &\textbf{24.0} &\textbf{41.0} &\textbf{44.6} &\textbf{47.1} &\textbf{18.8} &\textbf{36.9} \\ \bottomrule
\end{tabular}}
\caption{Ablation Studies of core components on MIT-States and C-GQA datasets.}
\label{tab:component}
\end{table*}

\begin{table*}[t]
\centering
    \begin{subtable}{0.5\linewidth}
    \centering
    \resizebox{0.8\textwidth}{!}{
    \setlength\tabcolsep{6pt}
    \renewcommand\arraystretch{0.9}
        \begin{tabular}{c|cccc}
        \toprule
        Expert dim $r$ & Unseen $\uparrow$  & Seen $\uparrow$  & AUC $\uparrow$  & HM $\uparrow$  \\ \cmidrule{1-5}
          $r=8$  & 42.0 & 45.9  & 16.8 & 34.7  \\  
          $r=16$  & 42.8 & 46.2  & 17.1 & 35.1  \\
          $r=32$ & 44.2 & 46.8 & 18.2 & 36.5   \\ 
          \rowcolor[rgb]{ .949, .949, .949} $r=64$  &\textbf{44.6} &\textbf{47.1} &\textbf{18.8} &\textbf{36.9}   \\ 
          $r=128$  & 43.8 & 46.5 &18.0 &36.1  \\ \bottomrule
        \end{tabular}}
    \setlength{\abovecaptionskip}{0.3cm}
    \setlength{\belowcaptionskip}{-0.0cm}
    \vspace{-5pt}
    \caption{\small{Hidden Dim of Expert Adapter}}
    \label{tab:dim}
    \end{subtable}
\hspace{-0.5em}     
    \begin{subtable}{0.5\linewidth}
    \centering
    \small
    \resizebox{0.8\textwidth}{!}{
    \setlength\tabcolsep{6pt}
    \renewcommand\arraystretch{0.9}
    \begin{tabular}{c|cccc}
    \toprule
    Number $K$ & Unseen $\uparrow$  & Seen $\uparrow$  & AUC $\uparrow$  & HM $\uparrow$  \\ \cmidrule{1-5}
        $K=0$  &42.4  &45.3 &16.7 &34.7  \\
        $K=1$   &43.8  &47.1 &18.0 &36.1 \\  
        \rowcolor[rgb]{ .949, .949, .949} $K=2$        &\textbf{44.6} &\textbf{47.1} &\textbf{18.8} &\textbf{36.9} \\ 
        $K=4$ &43.2  &46.9 &17.8 &35.9  \\ 
        $K=8$ &42.9  &46.4 &17.5 &35.7 \\ \bottomrule
    \end{tabular}
    }
    \setlength{\abovecaptionskip}{0.3cm}
    \setlength{\belowcaptionskip}{-0.0cm}
    \caption{\small{Activated Expert Number}}
    \label{tab:num}
    \end{subtable} \\
\vspace{+5px}
    \begin{subtable}{0.5\linewidth}
    \centering
    \resizebox{0.8\textwidth}{!}{
    \setlength\tabcolsep{6pt}
    \renewcommand\arraystretch{0.9}
        \begin{tabular}{c|cccc}
        \toprule
         Expert split & Unseen $\uparrow$  & Seen $\uparrow$  & AUC $\uparrow$  & HM $\uparrow$  \\ \cmidrule{1-5}
          $0+8$  & 43.5 & 46.5 &18.0 & 36.2  \\  
          \rowcolor[rgb]{ .949, .949, .949} $1+8$  &\textbf{44.6} &\textbf{47.1} &\textbf{18.8} &\textbf{36.9}  \\
          $2+8$ & 44.2 & 46.8 & 18.4 & 36.2   \\ 
          $4+4$ & 43.9 &46.2 &17.8 &35.9   \\ \bottomrule
        \end{tabular}}
    \setlength{\abovecaptionskip}{0.3cm}
    \setlength{\belowcaptionskip}{-0.0cm}
    \caption{\small{Expert Split}}
    \label{tab:share}
    \end{subtable}
\hspace{-0.5em}
    \begin{subtable}{0.5\linewidth}
    \centering
    \resizebox{0.85\textwidth}{!}{
    \setlength\tabcolsep{6pt}
    \renewcommand\arraystretch{0.9}
        \begin{tabular}{c|cccc}
        \toprule
        Variant Alignment & Unseen $\uparrow$  & Seen $\uparrow$  & AUC $\uparrow$  & HM $\uparrow$  \\ \cmidrule{1-5}
          \textsc{Baseline}  &42.8 &45.6  &17.2 &35.5  \\  
          + \textit{t2i alignment}  &43.8  &46.4  &18.0  &36.2  \\
          + \textit{inter-model affinity} &44.2&46.8&18.5&36.5   \\ 
          + \textit{intra-modal affinity}  &\textbf{44.6} &\textbf{47.1} &\textbf{18.8} &\textbf{36.9}   \\ \bottomrule
        \end{tabular}}
    \setlength{\abovecaptionskip}{0.3cm}
    \setlength{\belowcaptionskip}{-0.0cm}
    \caption{\small{Semantic Variant Alignment}}
    \label{tab:va}
    \end{subtable}
\vspace{-15pt}
\caption{Ablation experiments on C-GQA datasets.}
\vspace{-10pt}
\end{table*}

\noindent \textbf{Evaluation in Open-World Setting.} 
To evaluate the zero-shot performance of compositional generalization in real-world scenarios, we conduct comparative experiments in a more challenging open-world setting, where the target label space includes all possible state-object compositions. 
The results, presented in \cref{tab:open-world}, , demonstrate that our method significantly outperforms state-of-the-art approaches, achieving AUC gains of $+1.2\%$, $+2.9\%$ and $+1.2\%$ on MIT-States, UT-Zappos and C-GQA, respectively. 
Specifically, our method attains a notable Unseen score of $23.2\%$ on MIT-States dataset, surpassing the previous highest score of $21.8\%$. Additionally, our Seen score of $50.8\%$ exceeds that of several CZSL methods, highlighting a strong balance in handling both seen and unseen compositions. On UT-Zappos dataset, our method achieves an HM score of $54.2\%$, outperforming the previous state-of-the-art RAPR~\cite{li2024context} and demonstrating superior compositional generalization. On the most challenging C-GQA dataset, our method surpasses all other CZSL methods across all evaluation metrics, particularly in HM. 
The improvement of open-world performance demonstrates the effect of semantic variant alignment in establishing robust and accurate image-primitives matching relations.
Overall, our approach distinguishes itself through its ability to handle unseen data, making it highly effective for open-world tasks where new and unknown compositions are frequently encountered.

\begin{figure*}[tb]
\centering
\includegraphics[width=0.92\linewidth]{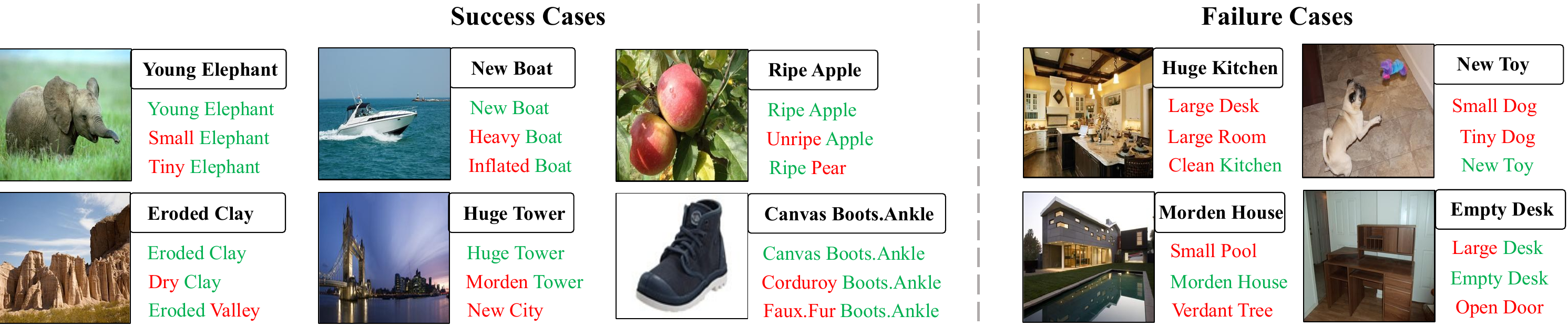}
\caption{Qualitative Results. We present top-3 predictions of randomly selected images in terms of success (Left) and failure cases (Right).}
\label{fig:img}
\end{figure*} 

\begin{figure*}[tb]
\centering
\includegraphics[width=\linewidth]{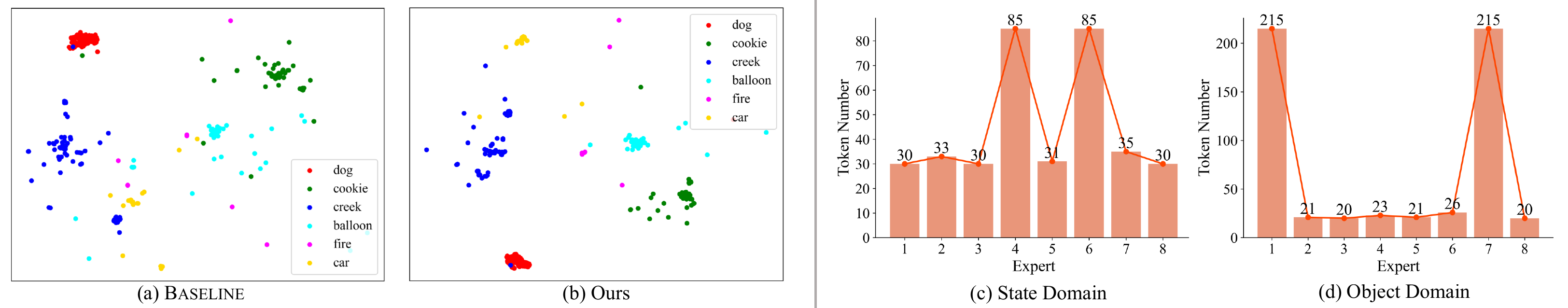}
\caption{\textbf{Left}: Visualization of image features learned by \textsc{Baseline} and our method. \textbf{Right}: The token load of various experts in state and object domains.}
\label{fig:fig}
\vspace{-15pt}
\end{figure*} 

\subsection{Ablation Study}
\noindent \textbf{Effect of Core Components.} We conduct several ablation experiments to access the effect of key components in our \textbf{EVA}, as presented in \cref{tab:component}. The \textsc{Baseline} employs the frozen CLIP~\cite{radford2021learning} encoder without the proposed methods, learning compositional zero-shot capacity with learnable prompts $P_s$, $P_o$ and $P_c$. \textit{Domain-expert adaption} boosts the model's zero-shot performance across all datasets, \eg, improving AUC from $20.2\%$ to $23.0\%$ on MIT-States, indicating that in-domain knowledge learning effectively strengthens primitive semantic modeling. Additionally, we observe that \textit{Semantic variant alignment} leads to a significant improvement, such as an AUC increase from $10.4\%$ to and $12.1\%$ on C-GQA. The integration of both components achieves a new state-of-the-art performance, further demonstrating the effectiveness of our approach.

\noindent \textbf{Expert dim $r$.} 
In \cref{tab:dim}, we investigate the influence of expert dim $r$ on compositional generalization performance. The optimal dimension is found to be 64, as it achieves the best performance while maintaining training efficiency. When the dimension is gradually reduced below 64, a decline in performance is observed, attributable to information loss caused by dimensional compression. Conversely, increasing the expert dimension to 128 does not yield performance improvements, suggesting that larger dimensions introduce information redundancy.

\noindent \textbf{Activated Expert Number.}
\cref{tab:num} reports the impact of activated expert number $K$. The optimal $K$ is 2, which achieves a win-win situation in terms of performance and computational cost. When we reduce $K$ to 0, \ie, only utilizing the sharing expert, the MoE adapter becomes a MLP-based adapter, resulting in a performance decline. We next set one activated expert and observe the improvement of CZSL performance, which suggests that the dynamic token routing is beneficial to primitives representation modeling. Larger number $K$ causes the lower performance, indicating the low efficiency in collaboration among multiple experts.

\noindent \textbf{Expert Split.}
We evaluate the impact of expert split, as depicted in \cref{tab:share}. By comparing configurations $0+8$ and $1+8$, we observe that shared expert enhances compositional recognition performance. However, increasing the number of shared experts leads to a decline in model performance, suggesting that retaining a certain proportion of routed experts is crucial. Furthermore, when the number of shared and routed experts is set to be equal, performance is lower than that of $0 + 8$. This indicates that dynamic routing plays a more significant role than constant collaboration in complex semantic learning.

\noindent \textbf{Semantic Variant Alignment.}
\cref{tab:va} provides ablation results for \textit{semantic variant alignment} on C-GQA dataset. Compared to \textsc{Baseline} without \textit{semantic variant alignment}, text-to-image (t2i) alignment effectively improves compositional zero-shot performance, \eg, $+0.8\%$ improvement in AUC. We observe that progressively incorporating the remaining modules (\ie, inter- and intra-modal affinity) further increases overall performance. The best results are achieved when all components are utilized, demonstrating the effectiveness of the proposed method.

\subsection{Qualitative Analysis}
\noindent \textbf{Qualitative Results.} 
As illustrated in \cref{fig:img}, we present the success (left) and failure (right) cases of top-3 predictions for randomly selected images. \textbf{EVA} demonstrates strong compositional recognition capabilities, accurately identifying complex semantic relationships between visual instances, such as \textit{Huge Tower} and \textit{Eroded Clay}. Furthermore, the top-3 predictions are semantically closely related to the Ground Truth, indicating that \textbf{EVA} establishes a robust and reliable cross-modal alignment for recognizing compositional relationships between states and objects. In failure cases, \textbf{EVA} produces incorrect yet semantically related predictions when interpreting holistic concepts (\textit{Huge Kitchen}) and ambiguous subjects (\textit{Modern House}). This highlights that a key focus for future compositional zero-shot learning (CZSL) research should be enhancing models' capacity for targeted and abstract understanding.

\noindent \textbf{Visualization of Feature Distributions.} \cref{fig:fig} (a)(b) visualizes the image features learned by \textsc{Baseline} and \textbf{EVA}. Leveraging \textit{domain-expert adaption} and \textit{semantic variant alignment}, \textbf{EVA} constructs a well-structured representation space, where features corresponding to identical states or objects are more tightly clustered, and class boundaries are more distinct. This structured representation enhances compositional generalization to unseen instances.

\noindent \textbf{Analysis on Expert Load.} To evaluate the impact of MoE adapter on in-domain knowledge learning, we analyze the computational load of each expert in learning state (\cref{fig:fig} (c)) and object (\cref{fig:fig} (d)). We observe an imbalanced distribution of token processing across experts in both state and object domains. In the state domain, experts $\mathcal{E}_4$ and $\mathcal{E}_6$
process the majority of tokens, while the remaining experts handle a similar token load. In the object domain, experts $\mathcal{E}_1$ and $\mathcal{E}_7$ exhibit the highest computational load. Since a single text encoder is utilized, this imbalance suggests knowledge separation, with certain experts specializing in specific domains, \ie, expert $\mathcal{E}_4$ excels in state-related tasks, while expert $\mathcal{E}_1$ performs well in object-related tasks.

%% file: sec/5_conclusion.tex
\vspace{-5pt}
\section{Conclusion}
In this work, we propose a Mixture-of-\underline{E}xpert Semantic \underline{V}ariant \underline{A}lignment framework (\textbf{EVA}) to address the challenges of concept learning and composition divergence within primitives. Inspired by distributed processing system of the human brain, we leverage MoE adapters to enable an end-to-end model without additional suffix modules. Through dynamic token allocation, experts specialize as effective in-domain learners, enhancing the modeling of primitive features. 
Moreover, we introduce \textit{semantic variant alignment} to enable fine-grained and accurate image-primitive mappings. The resulting well-structured primitive representation space facilitates the establishment of discriminative image-composition relations, improving compositional generalization. In future work, we aim to explore strategies to enhance the understanding of abstract concepts and the ability to distinguish object subjects effectively.

\noindent \textbf{Acknowledgements}
This work is supported by the National Natural Science Foundation of China through Grants 62088102, STI2030-Major Projects No.2022ZD0208801.

%% file: main.bbl
\begin{thebibliography}{42}
\providecommand{\natexlab}[1]{#1}
\providecommand{\url}[1]{\texttt{#1}}
\expandafter\ifx\csname urlstyle\endcsname\relax
  \providecommand{\doi}[1]{doi: #1}\else
  \providecommand{\doi}{doi: \begingroup \urlstyle{rm}\Url}\fi

\bibitem[Achiam et~al.(2023)Achiam, Adler, Agarwal, Ahmad, Akkaya, Aleman, Almeida, Altenschmidt, Altman, Anadkat, et~al.]{achiam2023gpt}
Josh Achiam, Steven Adler, Sandhini Agarwal, Lama Ahmad, Ilge Akkaya, Florencia~Leoni Aleman, Diogo Almeida, Janko Altenschmidt, Sam Altman, Shyamal Anadkat, et~al.
\newblock Gpt-4 technical report.
\newblock \emph{arXiv preprint arXiv:2303.08774}, 2023.

\bibitem[Atzmon et~al.(2016)Atzmon, Berant, Kezami, Globerson, and Chechik]{atzmon2016learning}
Yuval Atzmon, Jonathan Berant, Vahid Kezami, Amir Globerson, and Gal Chechik.
\newblock Learning to generalize to new compositions in image understanding.
\newblock \emph{arXiv preprint arXiv:1608.07639}, 2016.

\bibitem[Atzmon et~al.(2020)Atzmon, Kreuk, Shalit, and Chechik]{atzmon2020causal}
Yuval Atzmon, Felix Kreuk, Uri Shalit, and Gal Chechik.
\newblock A causal view of compositional zero-shot recognition.
\newblock \emph{Advances in Neural Information Processing Systems}, 33:\penalty0 1462--1473, 2020.

\bibitem[Bao et~al.(2024)Bao, Chen, Huang, and Kong]{bao2024prompting}
Wentao Bao, Lichang Chen, Heng Huang, and Yu Kong.
\newblock Prompting language-informed distribution for compositional zero-shot learning.
\newblock In \emph{European Conference on Computer Vision}, pages 107--123. Springer, 2024.

\bibitem[Brown et~al.(2020)Brown, Mann, Ryder, Subbiah, Kaplan, Dhariwal, Neelakantan, Shyam, Sastry, Askell, et~al.]{brown2020language}
Tom Brown, Benjamin Mann, Nick Ryder, Melanie Subbiah, Jared~D Kaplan, Prafulla Dhariwal, Arvind Neelakantan, Pranav Shyam, Girish Sastry, Amanda Askell, et~al.
\newblock Language models are few-shot learners.
\newblock \emph{Advances in neural information processing systems}, 33:\penalty0 1877--1901, 2020.

\bibitem[Devlin et~al.(2019)Devlin, Chang, Lee, and Toutanova]{Devlin2019BERTPO}
Jacob Devlin, Ming-Wei Chang, Kenton Lee, and Kristina Toutanova.
\newblock Bert: Pre-training of deep bidirectional transformers for language understanding.
\newblock In \emph{North American Chapter of the Association for Computational Linguistics}, 2019.

\bibitem[Dong et~al.(2019)Dong, Yang, Wang, Wei, Liu, Wang, Gao, Zhou, and Hon]{dong2019unified}
Li Dong, Nan Yang, Wenhui Wang, Furu Wei, Xiaodong Liu, Yu Wang, Jianfeng Gao, Ming Zhou, and Hsiao-Wuen Hon.
\newblock Unified language model pre-training for natural language understanding and generation.
\newblock \emph{Advances in neural information processing systems}, 32, 2019.

\bibitem[Hu et~al.(2021)Hu, Shen, Wallis, Allen-Zhu, Li, Wang, and Chen]{Hu2021LoRALA}
J.~Edward Hu, Yelong Shen, Phillip Wallis, Zeyuan Allen-Zhu, Yuanzhi Li, Shean Wang, and Weizhu Chen.
\newblock Lora: Low-rank adaptation of large language models.
\newblock \emph{ArXiv}, abs/2106.09685, 2021.

\bibitem[Huang et~al.(2024)Huang, Gong, Feng, Zhang, Lv, and Wang]{Huang_2024_CVPR}
Siteng Huang, Biao Gong, Yutong Feng, Min Zhang, Yiliang Lv, and Donglin Wang.
\newblock Troika: Multi-path cross-modal traction for compositional zero-shot learning.
\newblock In \emph{Proceedings of the IEEE/CVF Conference on Computer Vision and Pattern Recognition (CVPR)}, pages 24005--24014, 2024.

\bibitem[Isola et~al.(2015)Isola, Lim, and Adelson]{isola2015discovering}
Phillip Isola, Joseph~J Lim, and Edward~H Adelson.
\newblock Discovering states and transformations in image collections.
\newblock In \emph{Proceedings of the IEEE conference on computer vision and pattern recognition}, pages 1383--1391, 2015.

\bibitem[Jacobs et~al.(1991)Jacobs, Jordan, Nowlan, and Hinton]{Jacobs1991AdaptiveMO}
Robert~A. Jacobs, Michael~I. Jordan, Steven~J. Nowlan, and Geoffrey~E. Hinton.
\newblock Adaptive mixtures of local experts.
\newblock \emph{Neural Computation}, 3:\penalty0 79--87, 1991.

\bibitem[Jiang et~al.(2024)Jiang, Sablayrolles, Roux, Mensch, Savary, Bamford, Chaplot, de~las Casas, Hanna, Bressand, Lengyel, Bour, Lample, Lavaud, Saulnier, Lachaux, Stock, Subramanian, Yang, Antoniak, Scao, Gervet, Lavril, Wang, Lacroix, and Sayed]{jiang2024mixtralexperts}
Albert~Q. Jiang, Alexandre Sablayrolles, Antoine Roux, Arthur Mensch, Blanche Savary, Chris Bamford, Devendra~Singh Chaplot, Diego de~las Casas, Emma~Bou Hanna, Florian Bressand, Gianna Lengyel, Guillaume Bour, Guillaume Lample, Lélio~Renard Lavaud, Lucile Saulnier, Marie-Anne Lachaux, Pierre Stock, Sandeep Subramanian, Sophia Yang, Szymon Antoniak, Teven~Le Scao, Théophile Gervet, Thibaut Lavril, Thomas Wang, Timothée Lacroix, and William~El Sayed.
\newblock Mixtral of experts.
\newblock \emph{ArXiv}, abs/2401.04088, 2024.

\bibitem[Jing et~al.(2024)Jing, Li, Chen, and Shen]{jing2024retrieval}
Chenchen Jing, Yukun Li, Hao Chen, and Chunhua Shen.
\newblock Retrieval-augmented primitive representations for compositional zero-shot learning.
\newblock In \emph{Proceedings of the AAAI Conference on Artificial Intelligence}, pages 2652--2660, 2024.

\bibitem[Kingma and Ba(2014)]{kingma2014adam}
Diederik~P Kingma and Jimmy Ba.
\newblock Adam: A method for stochastic optimization.
\newblock \emph{arXiv preprint arXiv:1412.6980}, 2014.

\bibitem[Lake(2014)]{lake2014towards}
Brenden~M Lake.
\newblock \emph{Towards more human-like concept learning in machines: Compositionality, causality, and learning-to-learn}.
\newblock PhD thesis, Massachusetts Institute of Technology, 2014.

\bibitem[Lake et~al.(2017)Lake, Ullman, Tenenbaum, and Gershman]{lake2017building}
Brenden~M Lake, Tomer~D Ullman, Joshua~B Tenenbaum, and Samuel~J Gershman.
\newblock Building machines that learn and think like people.
\newblock \emph{Behavioral and brain sciences}, 40:\penalty0 e253, 2017.

\bibitem[Li et~al.(2021)Li, Selvaraju, Gotmare, Joty, Xiong, and Hoi]{li2021align}
Junnan Li, Ramprasaath Selvaraju, Akhilesh Gotmare, Shafiq Joty, Caiming Xiong, and Steven Chu~Hong Hoi.
\newblock Align before fuse: Vision and language representation learning with momentum distillation.
\newblock \emph{Advances in neural information processing systems}, 34:\penalty0 9694--9705, 2021.

\bibitem[Li et~al.(2022{\natexlab{a}})Li, Li, Xiong, and Hoi]{li2022blip}
Junnan Li, Dongxu Li, Caiming Xiong, and Steven Hoi.
\newblock Blip: Bootstrapping language-image pre-training for unified vision-language understanding and generation.
\newblock In \emph{International conference on machine learning}, pages 12888--12900. PMLR, 2022{\natexlab{a}}.

\bibitem[Li et~al.(2023)Li, Li, Savarese, and Hoi]{li2023blip}
Junnan Li, Dongxu Li, Silvio Savarese, and Steven Hoi.
\newblock Blip-2: Bootstrapping language-image pre-training with frozen image encoders and large language models.
\newblock In \emph{International conference on machine learning}, pages 19730--19742. PMLR, 2023.

\bibitem[Li et~al.(2022{\natexlab{b}})Li, Yang, Wei, Deng, and Yang]{li2022siamese}
Xiangyu Li, Xu Yang, Kun Wei, Cheng Deng, and Muli Yang.
\newblock Siamese contrastive embedding network for compositional zero-shot learning.
\newblock In \emph{Proceedings of the IEEE/CVF Conference on Computer Vision and Pattern Recognition}, pages 9326--9335, 2022{\natexlab{b}}.

\bibitem[Li et~al.(2024)Li, Liu, Chen, and Yao]{li2024context}
Yun Li, Zhe Liu, Hang Chen, and Lina Yao.
\newblock Context-based and diversity-driven specificity in compositional zero-shot learning.
\newblock In \emph{Proceedings of the IEEE/CVF Conference on Computer Vision and Pattern Recognition}, pages 17037--17046, 2024.

\bibitem[Li et~al.(2020)Li, Xu, Mao, and Lu]{li2020symmetry}
Yong-Lu Li, Yue Xu, Xiaohan Mao, and Cewu Lu.
\newblock Symmetry and group in attribute-object compositions.
\newblock In \emph{Proceedings of the IEEE/CVF Conference on Computer Vision and Pattern Recognition}, pages 11316--11325, 2020.

\bibitem[Liu et~al.(2024{\natexlab{a}})Liu, Feng, Xue, Wang, Wu, Lu, Zhao, Deng, Zhang, Ruan, et~al.]{liu2024deepseek}
Aixin Liu, Bei Feng, Bing Xue, Bingxuan Wang, Bochao Wu, Chengda Lu, Chenggang Zhao, Chengqi Deng, Chenyu Zhang, Chong Ruan, et~al.
\newblock Deepseek-v3 technical report.
\newblock \emph{arXiv preprint arXiv:2412.19437}, 2024{\natexlab{a}}.

\bibitem[Liu et~al.(2024{\natexlab{b}})Liu, Li, Wu, and Lee]{liu2024visual}
Haotian Liu, Chunyuan Li, Qingyang Wu, and Yong~Jae Lee.
\newblock Visual instruction tuning.
\newblock \emph{Advances in neural information processing systems}, 36, 2024{\natexlab{b}}.

\bibitem[Lu et~al.(2023)Lu, Guo, Liu, and Guo]{lu2023decomposed}
Xiaocheng Lu, Song Guo, Ziming Liu, and Jingcai Guo.
\newblock Decomposed soft prompt guided fusion enhancing for compositional zero-shot learning.
\newblock In \emph{Proceedings of the IEEE/CVF Conference on Computer Vision and Pattern Recognition}, pages 23560--23569, 2023.

\bibitem[Mancini et~al.(2021)Mancini, Naeem, Xian, and Akata]{mancini2021open}
Massimiliano Mancini, Muhammad~Ferjad Naeem, Yongqin Xian, and Zeynep Akata.
\newblock Open world compositional zero-shot learning.
\newblock In \emph{Proceedings of the IEEE/CVF conference on computer vision and pattern recognition}, pages 5222--5230, 2021.

\bibitem[Mancini et~al.(2022)Mancini, Naeem, Xian, and Akata]{mancini2022learning}
Massimiliano Mancini, Muhammad~Ferjad Naeem, Yongqin Xian, and Zeynep Akata.
\newblock Learning graph embeddings for open world compositional zero-shot learning.
\newblock \emph{IEEE Transactions on Pattern Analysis and Machine Intelligence}, 2022.

\bibitem[Mikolov(2013)]{mikolov2013efficient}
Tomas Mikolov.
\newblock Efficient estimation of word representations in vector space.
\newblock \emph{arXiv preprint arXiv:1301.3781}, 3781, 2013.

\bibitem[Misra et~al.(2017)Misra, Gupta, and Hebert]{misra2017red}
Ishan Misra, Abhinav Gupta, and Martial Hebert.
\newblock From red wine to red tomato: Composition with context.
\newblock In \emph{Proceedings of the IEEE Conference on Computer Vision and Pattern Recognition}, pages 1792--1801, 2017.

\bibitem[Naeem et~al.(2021)Naeem, Xian, Tombari, and Akata]{naeem2021learning}
Muhammad~Ferjad Naeem, Yongqin Xian, Federico Tombari, and Zeynep Akata.
\newblock Learning graph embeddings for compositional zero-shot learning.
\newblock In \emph{Proceedings of the IEEE/CVF Conference on Computer Vision and Pattern Recognition}, pages 953--962, 2021.

\bibitem[Nayak et~al.(2022)Nayak, Yu, and Bach]{nayak2022learning}
Nihal~V Nayak, Peilin Yu, and Stephen~H Bach.
\newblock Learning to compose soft prompts for compositional zero-shot learning.
\newblock \emph{arXiv preprint arXiv:2204.03574}, 2022.

\bibitem[Paszke et~al.(2019)Paszke, Gross, Massa, Lerer, Bradbury, Chanan, Killeen, Lin, Gimelshein, Antiga, et~al.]{paszke2019pytorch}
Adam Paszke, Sam Gross, Francisco Massa, Adam Lerer, James Bradbury, Gregory Chanan, Trevor Killeen, Zeming Lin, Natalia Gimelshein, Luca Antiga, et~al.
\newblock Pytorch: An imperative style, high-performance deep learning library.
\newblock \emph{Advances in neural information processing systems}, 32, 2019.

\bibitem[Purushwalkam et~al.(2019)Purushwalkam, Nickel, Gupta, and Ranzato]{purushwalkam2019task}
Senthil Purushwalkam, Maximilian Nickel, Abhinav Gupta, and Marc'Aurelio Ranzato.
\newblock Task-driven modular networks for zero-shot compositional learning.
\newblock In \emph{Proceedings of the IEEE/CVF International Conference on Computer Vision}, pages 3593--3602, 2019.

\bibitem[Radford et~al.(2019)Radford, Wu, Child, Luan, Amodei, Sutskever, et~al.]{radford2019language}
Alec Radford, Jeffrey Wu, Rewon Child, David Luan, Dario Amodei, Ilya Sutskever, et~al.
\newblock Language models are unsupervised multitask learners.
\newblock \emph{OpenAI blog}, 1\penalty0 (8):\penalty0 9, 2019.

\bibitem[Radford et~al.(2021)Radford, Kim, Hallacy, Ramesh, Goh, Agarwal, Sastry, Askell, Mishkin, Clark, et~al.]{radford2021learning}
Alec Radford, Jong~Wook Kim, Chris Hallacy, Aditya Ramesh, Gabriel Goh, Sandhini Agarwal, Girish Sastry, Amanda Askell, Pamela Mishkin, Jack Clark, et~al.
\newblock Learning transferable visual models from natural language supervision.
\newblock In \emph{International Conference on Machine Learning}, pages 8748--8763. PMLR, 2021.

\bibitem[Shazeer et~al.(2017)Shazeer, Mirhoseini, Maziarz, Davis, Le, Hinton, and Dean]{Shazeer2017OutrageouslyLN}
Noam~M. Shazeer, Azalia Mirhoseini, Krzysztof Maziarz, Andy Davis, Quoc~V. Le, Geoffrey~E. Hinton, and Jeff Dean.
\newblock Outrageously large neural networks: The sparsely-gated mixture-of-experts layer.
\newblock \emph{arXiv preprint arXiv:1701.06538}, 2017.

\bibitem[Team et~al.(2024)Team, Georgiev, Lei, Burnell, Bai, Gulati, Tanzer, Vincent, Pan, Wang, et~al.]{team2024gemini}
Gemini Team, Petko Georgiev, Ving~Ian Lei, Ryan Burnell, Libin Bai, Anmol Gulati, Garrett Tanzer, Damien Vincent, Zhufeng Pan, Shibo Wang, et~al.
\newblock Gemini 1.5: Unlocking multimodal understanding across millions of tokens of context.
\newblock \emph{arXiv preprint arXiv:2403.05530}, 2024.

\bibitem[Vaswani et~al.(2017)Vaswani, Shazeer, Parmar, Uszkoreit, Jones, Gomez, Kaiser, and Polosukhin]{vaswani2017attention}
Ashish Vaswani, Noam Shazeer, Niki Parmar, Jakob Uszkoreit, Llion Jones, Aidan~N Gomez, {\L}ukasz Kaiser, and Illia Polosukhin.
\newblock Attention is all you need.
\newblock \emph{Advances in neural information processing systems}, 30, 2017.

\bibitem[Xu et~al.(2022)Xu, Kordjamshidi, and Chai]{xu2022prompting}
Guangyue Xu, Parisa Kordjamshidi, and Joyce Chai.
\newblock Prompting large pre-trained vision-language models for compositional concept learning.
\newblock \emph{arXiv preprint arXiv:2211.05077}, 2022.

\bibitem[Xu et~al.(2024)Xu, Chai, and Kordjamshidi]{xu2024gipcol}
Guangyue Xu, Joyce Chai, and Parisa Kordjamshidi.
\newblock Gipcol: Graph-injected soft prompting for compositional zero-shot learning.
\newblock In \emph{Proceedings of the IEEE/CVF Winter Conference on Applications of Computer Vision}, pages 5774--5783, 2024.

\bibitem[Yu and Grauman(2014)]{yu2014fine}
Aron Yu and Kristen Grauman.
\newblock Fine-grained visual comparisons with local learning.
\newblock In \emph{Proceedings of the IEEE conference on computer vision and pattern recognition}, pages 192--199, 2014.

\bibitem[Zhou et~al.(2022)Zhou, Yang, Loy, and Liu]{zhou2022learning}
Kaiyang Zhou, Jingkang Yang, Chen~Change Loy, and Ziwei Liu.
\newblock Learning to prompt for vision-language models.
\newblock \emph{International Journal of Computer Vision}, 130\penalty0 (9):\penalty0 2337--2348, 2022.

\end{thebibliography}
